\RequirePackage{fix-cm}
\documentclass[smallextended]{svjour3}       
\smartqed  
\usepackage{graphicx,subfig}
\usepackage{amsmath}
\usepackage{multirow}
\begin{document}

\title{Prediction of laparoscopic procedure duration using unlabeled, multimodal sensor data}

\titlerunning{Online prediction of laparoscopic procedure duration}

\author{Sebastian Bodenstedt \and Martin Wagner \and
Lars M{\"u}ndermann \and Hannes Kenngott \and Beat M{\"u}ller-Stich \and Michael Breucha \and
S{\"o}ren Torge Mees \and J{\"u}rgen Weitz \and Stefanie Speidel}


\institute{S. Bodenstedt \and S. Speidel  \at
              Department for Translational Surgical Oncology, National Center for Tumor Diseases (NCT), Partner Site Dresden, Dresden, Germany \\
              \email{Firstname.Lastname@nct-dresden.de}           
           \and
           M. Wagner \and H. Kenngott \and B. M{\"u}ller-Stich\at
            Department of General, Visceral and Transplant Surgery, University of Heidelberg, Heidelberg
           \and
           L. M{\"u}ndermann \at
           KARL STORZ SE \& Co. KG, Tuttlingen, Germany
           \and
           M. Breucha \and S.T. Mees \and J. Weitz \at
           Department of Visceral, Thoracic and Vascular Surgery, Faculty of Medicine and University Hospital Carl Gustav Carus, TU Dresden, Dresden, Germany
}

\date{Received: date / Accepted: date}

\maketitle

\begin{abstract}
\textit{Purpose}
The course of surgical procedures is often unpredictable, making it difficult to estimate the duration of procedures beforehand.
This uncertainty makes scheduling surgical procedures a difficult task.
A context-aware method that analyses the workflow of an intervention online and automatically predicts the remaining duration would alleviate these problems.
As basis for such an estimate, information regarding the current state of the intervention is a requirement.

\textit{Methods}
Today, the operating room contains a diverse range of sensors.
During laparoscopic interventions, the endoscopic video stream is an ideal source of such information.
Extracting quantitative information from the video is challenging though, due to its high dimensionality.
Other surgical devices (e.g. insufflator, lights, etc.) provide data streams which are, in contrast to the video stream, more compact and easier to quantify.
Though whether such streams offer sufficient information for estimating the duration of surgery is uncertain.
In this paper, we propose and compare methods, based on convolutional neural networks, for continuously predicting the duration of laparoscopic interventions based on unlabeled data, such as from endoscopic image and surgical device streams.

\textit{Results}
The methods are evaluated on 80 recorded laparoscopic interventions of various types, for which surgical device data and the endoscopic video streams are available.
Here the combined method performs best with an overall average error of 37\% and an average halftime error of approximately 28\%.

\textit{Conclusion}
In this paper, we present, to our knowledge, the first approach for online procedure duration prediction using unlabeled endoscopic video data and surgical device data in a laparoscopic setting.
Furthermore, we show that a method incorporating both vision and device data performs better than methods based only on vision, while methods only based on tool usage and surgical device data perform poorly, showing the importance of the visual channel.
\end{abstract}

\section{Introduction}
The time in the operating room (OR) and the time of the operating staff are cost intensive hospital resources and have to be allocated precisely.
Planning the usage of the OR cannot be static, as a procedure that takes longer than previously estimated can cause the following surgeries to be pushed back or even canceled, thereby inconveniencing both patient and the surgical team.
On the other hand, if a procedure finishes early, the OR stays unused, incurring unnecessary idle time for the surgical personnel.
To prevent this from occurring, OR schedulers need to dynamically update timetables.
This is complicated by the unpredictability of the surgical workflow, which makes it difficult to estimate the duration of procedures beforehand.
Therefore, the OR schedulers have to be constantly kept in the loop of the progress of ongoing interventions.
For this, they have to periodically inquire into the status of interventions, resulting in highly subjective predictions and avoidable interruptions of procedures.

Integrated ORs are becoming more prevalent in hospitals, making it possible to access data streams from surgical devices such as cameras, insufflator, lights, etc. during interventions.
Such data streams can provide information that enable context-aware assistance, such as automatically and continuously predicting the progress of an ongoing intervention.
Especially the endoscopic video stream, via which laparoscopic interventions are performed, contains a large supply of information.
Workflow analysis methods can be used to segment interventions into surgical phases.
Often tool usage is employed to determine the current surgical phase \cite{Blum2010,Dergachyova2016,Katic2014,Padoy2012632}, which provides an indicator for progress surgery.
Convolutional neural networks (CNNs) have also been used to determine the surgical phase directly from the endoscopic video stream \cite{Bodenstedt2017,Lea2016,TwinandaSMMMP16}.

Surgical phase detection methods can be used to approximate the duration of surgical procedures, but these methods generally require a sufficient amount of labeled examples as training input.
Furthermore, seeing that phase models are generally specified to a certain type of intervention, multiple detectors would need to be trained.
Therefore, using a phase based method as a general solution to determine the remaining duration of surgeries would require an unfeasible large amount of labeled training data.
In \cite{Guedon2016}, the authors propose a system that determines the remaining time of surgery during laparoscopic cholecystectomies without surgical phases, but directly from the usage of the electrosurgical device.
Two recurrent CNNs for predicting the remaining time of surgery directly from endoscopic video of cholecystectomies are presented in \cite{aksamentov2017deep} and in \cite{twinanda2018rsdnet}.
While \cite{twinanda2018rsdnet} also does not rely on the annotation of surgical phases, only visual features are used as input.
In \cite{Bodenstedt2018}, we also propose a method for computing the duration of laparoscopic surgeries of varying types using a combination of a CNN and a gated recurrent unit \cite{ChoMBB14} (GRU) with unlabeled video sequences.

In this paper, we propose and evaluate three methods, based on recurrent CNNs, for directly predicting and refining the duration of laparoscopic interventions.
These methods do not require labeled training data and function for different types of laparoscopic procedures.
The first method uses surgical device data collected from the OR as input, the second endoscopic image data. 
The third method is the combination of the previous methods.
The evaluation of the methods is performed on a dataset containing 80 laparoscopic surgeries of varying types and on the Cholec80 dataset \cite{TwinandaSMMMP16}.
To our knowledge, our approach is the first method to predict the duration of laparoscopic interventions based on a combination of unlabeled vision and surgical device data.
\section{Methods}
A requirement for predicting the remaining duration of laparoscopic procedures is information regarding the current state of the surgery.
As the endoscopic video stream serves as basis for the actions of the surgeon, we assume that it contains sufficient information on the state of the procedure, though extracting quantitative information is challenging due to the high dimensionality of the data stream.
Here, we propose a recurrent CNN that allows predicting the duration of surgery from an endoscopic video stream.
Furthermore, we propose a variation of this recurrent CNN that explicitly performs tool presence detection and utilizes this information to further enhance its predictions.

\begin{figure}[tb]
\centering
   \includegraphics[width=0.75\columnwidth]{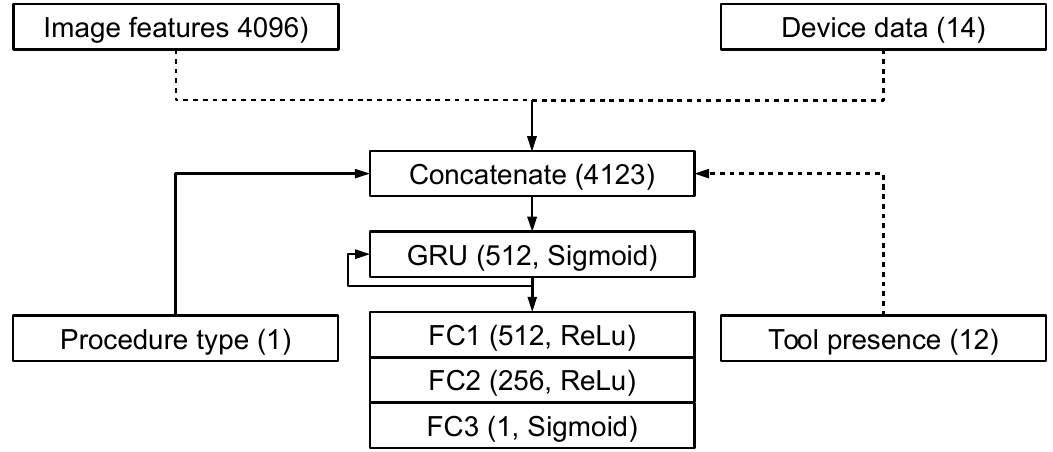}
   \caption{The recurrent CNN topology for predicting surgery duration from images, device data and tool presence. The dotted lines indicate optional inputs.
	\textit{FC} stands for a fully connected layer and \textit{GRU} for a gated recurrent unit. 
	The values in parenthesis indicate the number of hidden units and the nonlinearity used.}
   \label{fig:cnn}       
\end{figure}

On the other hand, integrated ORs are starting to become more prevalent in hospitals.
These ORs make it possible to access data streams, in the form of time series, from other surgical devices.
These time series are, in contrast to the video stream, more compact and easier to quantify, but contain a smaller amount of information.
We hypothesize that both streams, video and device data, contain complementary information, meaning that combining the two should increase prediction accuracy.
To evaluate this claim, we propose a fusion of the two streams into one recurrent CNN (figure \ref{fig:cnn}).
\subsection{V-Net: Vision based estimation}
Building upon our work in \cite{Bodenstedt2018}, the topology of our recurrent CNN for predicting duration of surgery from the endoscopic video, V-Net, is identical to the network for phase segmentation proposed in \cite{Bodenstedt2017}, which consists of an Alexnet \cite{Alexnet} style network (image features in figure \ref{fig:cnn}) combined with a GRU to incorporate temporal information.
Only the final layers differs, we use a fully-connected layer with a single output and a sigmoid activation function.
We pretrain the layers preceding the GRU with the method proposed in \cite{Bodenstedt2017}.
For this, we extract 2.2 million frames from 324 unlabeled laparoscopic procedures.
The Alexnet is then trained in a Siamese fashion to extract features from images that allow sorting two random, given frames from the same procedure into the correct temporal order.

As input for V-Net, we sampled videos at a rate of one frame per second and downsampled each image to a resolution of $320\times240$ pixels.
This is performed to reduce data size and training time.

Furthermore, we provide V-Net with information on the type of laparoscopic surgery being performed (procedure type in figure \ref{fig:cnn}).
For this, we categorize our dataset into 5 general types of laparoscopic surgery (table \ref{tab:types_surgery}).
We assign each frame a number between 0 and 1 as label, i.e. the label of frame $i$ from a video consisting of $N$ frames is $l_i = \frac{i}{N}$.
During inference, we can then directly compute the duration prediction $\tilde{N}_i$:
$\tilde{N}_i = \frac{i}{y_i}$, where $y_i \in [0,1]$ is the predicted progress of the procedure.
\begin{table}[hbt]
\centering
\begin{tabular}{|l|l|l|l|}
\hline
ID&Surgery type&Samples in dataset&Average length (min.)\\
\hline
1&Colorectal&39&156\\
\hline
2&Upper Gastrointestinal and Bariatric&11&107\\
\hline
3&Hepato-Pancreatico-Biliary&4&102\\
\hline
4&General Laparoscopic &21&41\\
\hline
5&Singular case&5&91\\
\hline
\end{tabular}
\caption{The categorization of laparoscopic surgery used as input. Also shown, the number of occurrences and average length in the \textbf{MultiType} dataset.}
\label{tab:types_surgery}
\end{table}

Since the layers preceding the GRU are pretrained, we only optimize the weights of the newly added layers. 
For this, we use Adam \cite{kingma2014adam} with a learning rate of $10^{-6}$ and a sigmoid cross-entropy loss.
The network is trained for 50 epochs.
\subsection{T-Net: Tool presence based estimation}
Seeing that surgical tool presence in the endoscopic video is an important cue for progress of surgery \cite{TwinandaSMMMP16}, we hypothesize that information on the types of tools currently visible could be used to predict procedure duration.
Tool usage could in principle be determined through external sensors such as RFID tags or a monitoring system.
To simulate such a system and to identify the tools currently in use in the endoscopic video stream, we decided to apply a CNN trained on weakly labeled data were only tool usage is annotated in form of binary labels.
Such data is easily annotated and is generally not specific to one type of surgical procedure.

For this, we modified a pretrained ResNet-152 \cite{he2016deep} for tool presence detection.
Here, we replace the pretrained fully connected output layer with another fully connected layer, consisting of 12 nodes (one per tool type) with sigmoid nonlinearities.
The weights of the \textit{conv5\_x} layers and of the new fully connected layer are fine-tuned using a dataset of 24 colorectal laparoscopies in which the presence of 12 possible surgical tools has been labeled in one frame per minute.
We used every labeled frame in the dataset to train the ResNet-152, overall 3800 frames.
Adam \cite{kingma2014adam} with a learning rate of $10^{-5}$ and binary cross-entropy are used.
The network was then tested on a separate set of 9 further colorectal laparoscopies, consisting of 1543 annotated video frames.
An average F1-score over all class of 81.2\% was achieved.

To use tool presence information for predicting procedure duration, we use the output of the ResNet152 directly as input for the GRU described in the previous section, replacing the image features (tool presence in figure \ref{fig:cnn}).
T-Net is trained with the same hyper-parameters as V-Net.
\subsection{D-Net: Surgical device data based estimation}
The input consists of 14 values, each representing a different signal from a surgical device (a list of the devices and signals used can be found in table \ref{tab:devices}).
Analog to the video stream, we select one second as the size of each time step. 
For signals with a data rate higher than 1 Hz, we discard all values, except the most recent one.
For signals with a lower data rate, we use the most recent value, even if it was older than one second.

To predict the remaining duration of surgery, we first normalize the data, given the value ranges provided in table \ref{tab:devices}.
The data is then fed into the previously described GRU (device data in figure \ref{fig:cnn}).
D-Net is trained with the same hyper-parameters as the previous networks.

\subsection{TD-Net: Tool presence and surgical device data based estimation}
To utilize tool presence information and surgical data for predicting the duration of surgery, we combine D-Net and T-Net into one architecture.
This ID-Net is trained with identical hyper-parameters as the previous networks.
\subsection{VT-Net: Vision and tool presence based estimation}

To combine the tool presence information for predicting duration of surgery with the image features, we extend the architecture of V-Net into VT-Net, allowing it to additionally accept information on tool presence.
VT-Net is trained with the same hyper-parameters as the previous networks.

\subsection{VTD-Net: incorporating surgical device data}
To incorporate the data stream provided by surgical devices, we extend VT-Net with a further input.
VTD-Net is trained with the same hyper-parameters as the previous networks.

\begin{table}[hbt]
\centering
\begin{tabular}{|l|l|l|l|}
\hline
Device&Signal&Data type&Value range\\
\hline
\multirow{7}{*}{Insufflator}& Current gas flow rate&Continuous&0-215\\
\cline{2-4}
& Target gas flow rate&Continuous&10-300\\
\cline{2-4}
& Current gas pressure&Continuous&0-255\\
\cline{2-4}
& Target gas pressure&Continuous&9-23\\
\cline{2-4}
& Used gas volume&Continuous&0-9501\\
\cline{2-4}
& Gas supply pressure&Continuous&0-760\\
\cline{2-4}
& Device on?&Binary&0,1\\
\hline
\multirow{3}{*}{OR lights}& All lights off?&Binary&0,1\\
\cline{2-4}
& Intensity light 1&Continuous&0-100\\
\cline{2-4}
& Intensity light 2&Continuous&0-100\\
\hline
Endoscopic light source & Intensity&Continuous&0-100\\
\hline
\multirow{3}{*}{Endoscopic camera}& White balance&Binary&0,1\\
\cline{2-4}
& Gains&Continuous&0-3298\\
\cline{2-4}
& Exposure index&Continuous&0-834\\
\hline
\end{tabular}
\caption{Surgical devices and signals used}
\label{tab:devices}
\end{table}

\section{Experiments and results}
The basis of our evaluation is a dataset, containing recordings of 80 laparoscopic surgeries of different procedure types (\textbf{MultiType}).
For each surgery, the dataset contains the endoscopic video stream and data collected from different surgical devices as listed in table \ref{tab:devices}.
The procedures were all recorded in the same OR using the integrated operating room system OR1\texttrademark\ (Karl Storz GmbH \& Co KG, Tuttlingen, Germany).
The average procedure length in the dataset is 106 minutes.
The datasets used for pretraining V-Net and training the ResNet-152 for tool presence detection do not overlap with this dataset.

To evaluate the proposed methods, we divide the dataset into four sets of equal size and perform a leave-one-set-out evaluation for each method.
While dividing the dataset, we ascertain that the distribution of the different types of surgery into each set is balanced.

During testing, we compute the duration prediction $\tilde{N}_i$ at each frame $i$:
$\tilde{N}_i = \frac{i}{y_i}$, where $y_i \in [0,1]$ is the predicted progress of the procedure.
With $\tilde{N}_i$, we can compute the absolute duration prediction error in seconds, $|\tilde{N}_i - N|$, and the duration prediction error relative to the length $N$ of of each procedure,
$\frac{|\tilde{N}_i - N|}{N}$.
The relative error gives a more appropriate impression on how well each method can predict procedure duration.

To put the performance of the proposed methods into perspective, we also propose different baselines for computing the remaining duration of a given surgery.
\paragraph{Naive}
As a \textbf{naive} baseline, we provide the duration prediction error that would occur if the average procedure duration over the training data were used as value for the predicted progress of the procedure.
\paragraph{Type}
We also provide a \textbf{type}-based baseline, where we instead compute the average procedure duration separately for each procedure category in table \ref{tab:types_surgery}.
\paragraph{Twinanda et al.}
As a further baseline, we implemented the method proposed by Twinanda et al. \cite{twinanda2018rsdnet} for predicting the remaining surgery duration.
The method uses a ResNet152 to extract features from the images, which are then processed by a long short-term memory unit (LSTM) \cite{hochreiter1997long} to predict the remaining surgery time in minutes.
As the operations in \textbf{MultiType} are significantly longer than the cholecystectomies in the dataset in \cite{twinanda2018rsdnet}, we set the parameter $s_{norm}$ to 20.
Furthermore, the paper did not specify how many units the LSTM contained, we opted to use 512 units.
Also the image size was not mentioned, we therefore used the default input resolution of ResNet152, which is $224\times224$ pixels.
For training and testing \textbf{MultiType}, we performed a leave-one-set-out evaluation, training the ResNet152 and the LSTM in two steps as proposed in \cite{twinanda2018rsdnet}, though all the training data was used for both CNN and LSTM.
We used 60000 iterations to train the LSTM.
All other parameter values were set to equal values as described in \cite{twinanda2018rsdnet}.

For each method, we provide both the absolute and the relative average duration prediction error (see tables \ref{tab:results:abs} and \ref{tab:results:rel}).
To measure how the error progresses during the course of a procedure, we compute the average error during each quarter of a given procedure (Q1-Q4).

\begin{table}[hbt]
\centering
\subfloat[Absolute error (in seconds)]{
\label{tab:results:abs}
\begin{tabular}{|l|l|l|l|l||l|}
\hline
Method&Q1&Q2&Q3&Q4&Mean\\
\hline
V-Net&$3818\pm486$&$2496\pm258$&$1611\pm230$&$1353\pm200$&$2320\pm846$\\
\hline
T-Net&$5078\pm1004$&$1611\pm991$&$1857\pm1000$&$4316\pm1006$&$3215\pm2005$\\
\hline
D-Net&$4406\pm1315$&$1928\pm788$&$2301\pm910$&$4482\pm1667$&$3280\pm2373$\\
\hline
ID-Net&$4473\pm1330$&$1835\pm815$&$2181\pm912$&$4191\pm1613$&$3170\pm2288$\\
\hline
VT-Net&$4271\pm433$&$2421\pm318$&$1052\pm285$&$1457\pm322$&$2300\pm886$\\
\hline
VTD-Net&$4143\pm449$&$2289\pm312$&$1071\pm279$&$1313\pm312$&$2204\pm875$\\
\hline
\hline
Baseline (Naive)&$3208\pm2085$&$3208\pm2085$&$3208\pm2085$&$3208\pm2085$&$3208\pm2085$\\
\hline
Baseline (Type)&$2093\pm1787$&$2093\pm1787$&$2093\pm1787$&$2093\pm1787$&$2093\pm1787$\\
\hline
Twinanda et al.&$3862\pm531$&$2427\pm508$&$1405\pm463$&$1828\pm507$&$2380\pm1480$\\
\hline
\end{tabular}}

\subfloat[Relative error]{
\label{tab:results:rel}
\begin{tabular}{|l|l|l|l|l||l|}
\hline
Method&Q1&Q2&Q3&Q4&Mean\\
\hline
V-Net&$53\%\pm9\%$&$42\%\pm6\%$&$42\%\pm5\%$&$44\%\pm4\%$&$45\%\pm16\%$\\
\hline
T-Net&$74\%\pm15\%$&$23\%\pm15\%$&$28\%\pm15\%$&$79\%\pm15\%$&$51\%\pm30\%$\\
\hline
D-Net&$65\%\pm20\%$&$28\%\pm13\%$&$45\%\pm15\%$&$80\%\pm22\%$&$54\%\pm37\%$\\
\hline
TD-Net&$66\%\pm20\%$&$27\%\pm13\%$&$42\%\pm16\%$&$75\%\pm22\%$&$53\%\pm36\%$\\
\hline
VT-Net&$61\%\pm8\%$&$35\%\pm6\%$&$25\%\pm5\%$&$37\%\pm6\%$&$39\%\pm15\%$\\
\hline
VTD-Net&$59\%\pm8\%$&$32\%\pm5\%$&$24\%\pm5\%$&$35\%\pm6\%$&$37\%\pm14\%$\\
\hline
\hline
Baseline (Naive)&$124\%\pm245\%$&$124\%\pm245\%$&$124\%\pm245\%$&$124\%\pm245\%$&$124\%\pm245\%$\\
\hline
Baseline (Type)&$57\%\pm99\%$&$57\%\pm99\%$&$57\%\pm99\%$&$57\%\pm99\%$&$57\%\pm99\%$\\
\hline
Twinanda et al.&$65\%\pm9\%$&$54\%\pm8\%$&$51\%\pm8\%$&$66\%\pm9\%$&$59\%\pm23\%$\\
\hline
\end{tabular}}
\caption{The average duration prediction errors on the \textbf{MultiType} dataset. \protect\subref{tab:results:abs} shows the mean absolute error of all the methods and \protect\subref{tab:results:rel} the mean relative error for all the methods.}
\label{tab:results:HD}
\end{table}

For further evaluation, V-Net, T-Net and VT-Net are applied to the publicly available \textbf{Cholec80} dataset \cite{TwinandaSMMMP16}, which consists of 80 videos from laparoscopic cholecystectomies with an average length of 38 minutes.
We divide the dataset into four sets of equal size and similar average procedure length and perform four leave-one-set-out evaluations for each method.
The absolute and the relative average duration prediction errors can be found in tables \ref{tab:results_ch:abs} and \ref{tab:results_ch:rel}.
The type-based baseline is not available, as the dataset consists only of a single procedure type.
For the baseline based on Twinanda et al., all parameters were set to identical values as in \cite{twinanda2018rsdnet}, though we again used all the data in the training set to train both the ResNet152 and the LSTM.
No surgical device data is contained in the dataset, meaning the networks relying on surgical device data cannot be used.
\begin{table}[hbt]
\centering
\subfloat[Absolute error (in seconds)]{
\label{tab:results_ch:abs}
\begin{tabular}{|l|l|l|l|l||l|}
\hline
Method&Q1&Q2&Q3&Q4&Mean\\
\hline
V-Net&$934\pm238$&$804\pm135$&$581\pm119$&$353\pm92$&$668\pm233$\\
\hline
T-Net&$1700\pm350$&$500\pm333$&$722\pm350$&$735\pm351$&$915\pm705$\\
\hline
VT-Net&$920\pm147$&$746\pm104$&$434\pm93$&$236\pm80$&$584\pm214$\\
\hline
\hline
Baseline (Naive)&$768\pm669$&$768\pm669$&$768\pm669$&$768\pm669$&$768\pm669$\\
\hline
Twinanda et al.&$932\pm284$&$619\pm208$&$478\pm211$&$324\pm217$&$588\pm402$\\
\hline
\end{tabular}}

\subfloat[Relative error]{
\label{tab:results_ch:rel}
\begin{tabular}{|l|l|l|l|l||l|}
\hline
Method&Q1&Q2&Q3&Q4&Mean\\
\hline
V-Net&$42\%\pm12\%$&$38\%\pm7\%$&$31\%\pm6\%$&$22\%\pm5\%$&$33\%\pm11\%$\\
\hline
T-Net&$74\%\pm15\%$&$22\%\pm14\%$&$31\%\pm15\%$&$35\%\pm15\%$&$41\%\pm31\%$\\
\hline
VT-Net&$34\%\pm7\%$&$28\%\pm5\%$&$17\%\pm4\%$&$12\%\pm4\%$&$23\%\pm9\%$\\
\hline
\hline
Baseline (Naive)&$66\%\pm27\%$&$66\%\pm27\%$&$66\%\pm27\%$&$66\%\pm27\%$&$66\%\pm27\%$\\
\hline
Twinanda et al.&$36\%\pm13\%$&$27\%\pm9\%$&$23\%\pm11\%$&$17\%\pm11\%$&$25\%\pm17\%$\\
\hline
\end{tabular}}
\caption{The average duration prediction errors on the \textbf{Cholec80} dataset. \protect\subref{tab:results_ch:abs} shows the mean absolute error of all the methods and \protect\subref{tab:results_ch:rel} the mean relative error for all the methods.}
\label{tab:results:cholec}
\end{table}

The results on \textbf{MultiType} show that all three of our methods that are based on the image features, outperform the baseline methods.
All the image feature based methods also outperform the baseline based on Twinanda et al., also it can be noted that the methods presented here show a lower standard deviation of the error.
One explanation why even V-Net performs better than Twinanda et al. could be due to the pretraining that was performed.
Furthermore, VTD-Net provides more accurate results than the other two image feature based networks, demonstrating that surgical device data does indeed contain complementary information.
Table \ref{tab:results_type:HD} shows that VTD-Net performs consistently on the first three procedure categories.
As can be seen in table \ref{tab:types_surgery}, the surgeries in the first category are significantly longer than those in the second and third categories, though the relative error seems to stay consistent.
Part of the general laparoscopic category are diagnostic laparoscopies, which are significantly shorter ($< 15$min) than the average procedure in the dataset and are less standardized and therefore difficult to predict, explaining this drop in performance.
The last category contains only singular cases, which differ from the other categories, making predictions difficult.

The non-image based methods generally perform worse than the image based methods, leading us to conclude that while these data sources contain hints relevant for the progress of surgery that are useful for augmenting the image features, by themselves they don't contain sufficient information to accurately predict the progress of surgery. 

On \textbf{Cholec80}, both image feature based methods outperform the naive baseline, while VT-Net also outperformed V-Net.
The baseline based on Twinanda et al. outperformed V-Net and performed similarly to VT-Net, thought the standard deviation of the error of both V-Net and VT-Net was smaller.
The method of Twinanda et al. seems to perform better at the beginning of surgery, while VT-Net performs better in the second half of surgery.
Similarly as for \textbf{MultiType}, the method solemnly based on tool usage performs poorly. 
A direct comparison with the results in \cite{aksamentov2017deep} is not possible, as the authors use a private dataset for testing.
\begin{table}[hbt]
\centering
\subfloat[Absolute error (in seconds)]{
\label{tab:results_type:abs}
\begin{tabular}{|l|l|l|l|l||l|}
\hline
Type&Q1&Q2&Q3&Q4&Mean\\
\hline
1&$6020\pm961$&$3452\pm733$&$1350\pm652$&$1381\pm710$&$3050\pm2136$\\
\hline
2&$4102\pm863$&$2096\pm541$&$852\pm474$&$1226\pm527$&$2069\pm1460$\\
\hline
3&$4070\pm822$&$2090\pm510$&$764\pm376$&$1053\pm475$&$1994\pm1483$\\
\hline
4&$1098\pm457$&$496\pm241$&$706\pm246$&$1270\pm292$&$893\pm572$\\
\hline
5&$1419\pm522$&$707\pm308$&$927\pm256$&$1518\pm382$&$1143\pm746$\\
\hline
\end{tabular}}

\subfloat[Relative error]{
\label{tab:results_type:rel}
\begin{tabular}{|l|l|l|l|l||l|}
\hline
Type&Q1&Q2&Q3&Q4&Mean\\
\hline
1&$67\%\pm12\%$&$37\%\pm8\%$&$15\%\pm7\%$&$18\%\pm8\%$&$34\%\pm24\%$\\
\hline
2&$62\%\pm14\%$&$30\%\pm9\%$&$12\%\pm7\%$&$22\%\pm9\%$&$32\%\pm22\%$\\
\hline
3&$65\%\pm14\%$&$32\%\pm8\%$&$12\%\pm6\%$&$18\%\pm8\%$&$32\%\pm24\%$\\
\hline
4&$43\%\pm20\%$&$25\%\pm12\%$&$45\%\pm13\%$&$72\%\pm14\%$&$46\%\pm28\%$\\
\hline
5&$45\%\pm21\%$&$26\%\pm13\%$&$46\%\pm11\%$&$71\%\pm15\%$&$47\%\pm30\%$\\
\hline
\end{tabular}}
\caption{The average duration prediction errors of VTD-Net on the \textbf{MultiType} dataset, broken down according to procedure type. \protect\subref{tab:results_type:abs} shows the mean absolute error and \protect\subref{tab:results_type:rel} the mean relative error.}
\label{tab:results_type:HD}
\end{table}

\section{Discussion}
In this paper, we presented, to our knowledge, the first approach for online procedure duration prediction using unlabeled endoscopic video data and surgical device data in a laparoscopic setting.
On \textbf{MultiType}, VTD-Net showed an overall average prediction error of 37\% and a halftime error of about 28\%, which are lower than the results from the baselines. 
Furthermore, we showed that a method incorporating both vision and device data performs better than methods based only on vision, while methods only based on tool usage and surgical device data performed poorly, showing the importance of the visual channel.
The evaluation showed that the presented methods currently produce larger than average errors on irregular procedures with short length (shorter than 15 min) and on singular cases.
As the methods performed better on \textbf{Cholec80}, which contains mostly shorter and more standardized operations, we assume this is due to a lack of training data of the irregular cases.
When looking at complex procedures of medium length (second and third category) and long length (first category), the relative error stays consistent.

On \textbf{Cholec80}, VT-Net achieved an average prediction error as well as a halftime error of 23\%, outperforming the baseline.
This difference of performance on the two datasets can be contributed to the fact that \textbf{MultiType} contained multiple procedure types and also had a higher variance in procedure duration.
Seeing that the results of the proposed methods perform significantly better on a simple operation like cholecystectomies than on a more complex and diverse operations as in the \textbf{MultiType} dataset leads us to conclude that the proposed methods themselves are sound, but that more training data is required for more complex cases.

As our results indicate that combing vision and device data provides more information on the progress of surgery, we assume that data available to the anesthetist, such as heart rate, blood pressure and drug doses, would provide even more valuable insights.

\small{
\textbf{Conflict of interest}
S. Bodenstedt, M. Wagner, L. M{\"u}ndermann, H. Kenngott, B. M{\"u}ller-Stich,  M. Breucha, S. Mees, J. Weitz and S. Speidel declare that they have no conflict of interest.

\textbf{Ethical approval}
All procedures performed in studies involving human participants were in accordance with the ethical standards of the institutional and/or national research committee and with the 1964 Declaration of Helsinki and its later amendments or comparable ethical standards

\textbf{Informed consent}
Informed consent was obtained from the study participants
}
\bibliographystyle{spmpsci}      
\bibliography{paper}   

\end{document}